\documentclass{article}
\usepackage{ijcai25}

\usepackage{amsthm}
\usepackage{times}
\usepackage{soul}
\usepackage{url}
\usepackage[hidelinks]{hyperref}
\usepackage[utf8]{inputenc}
\usepackage[small]{caption}
\usepackage{graphicx}
\usepackage{subcaption}
\usepackage{amsmath}
\usepackage{thm-restate}
\usepackage{booktabs}
\usepackage{algorithm}
\usepackage{graphicx}
\usepackage[table]{xcolor}
\definecolor{lightgray}{gray}{0.9}
\usepackage{algpseudocode}
\usepackage[switch]{lineno}
\usepackage{amssymb}
\usepackage{multirow}
\usepackage{float}
\usepackage{mathtools}
\usepackage{thmtools, thm-restate}
\theoremstyle{plain}
\newtheorem{definition}{Definition}

\usepackage{booktabs}
\usepackage[table]{xcolor} 
\definecolor{lightgray}{gray}{0.85}
\definecolor{lightblue}{RGB}{170, 220,224}
\definecolor{lightorange}{RGB}{255,175,176}

\title{Zero-Shot Machine Unlearning with Proxy Adversarial Data Generation}

\author{
Huiqiang Chen$^{1,2}$
\footnote{work done while at University of Queensland}
\and
Tianqing Zhu$^{1}$ \footnote{corresponding author}
\and
Xin Yu$^{3}$
\and
Wanlei Zhou$^1$\\
\affiliations
$^1$City University of Macau, Macau, China\\
$^2$University of Technology Sydney, NSW, Australia\\
$^3$University of Queensland, QLD, Australia\\
\emails
cs.hqchen@gmail.com,
\{tqzhu, wlzhou\}@cityu.edu.mo,
xin.yu@uq.edu.au
}

\begin{document}

\maketitle

\begin{abstract}
Machine unlearning aims to remove the influence of specific samples from a trained model. A key challenge in this process is over-unlearning, where the model's performance on the remaining data significantly drops due to the change in the model's parameters. Existing unlearning algorithms depend on the remaining data to prevent this issue. As such, these methods are inapplicable in a more practical scenario, where only the unlearning samples are available (i.e., zero-shot unlearning). This paper presents a novel framework, ZS-PAG, to fill this gap. Our approach offers three key innovations: (1) we approximate the inaccessible remaining data by generating adversarial samples; (2) leveraging the generated samples, we pinpoint a specific subspace to perform the unlearning process, therefore preventing over-unlearning in the challenging zero-shot scenario; and (3) we consider the influence of the unlearning process on the remaining samples and design an influence-based pseudo-labeling strategy. As a result, our method further improves the model's performance after unlearning. The proposed method holds a theoretical guarantee, and experiments on various benchmarks validate the effectiveness and superiority of our proposed method over several baselines. 
\end{abstract}

\section{Introduction} The success of modern AI systems relies heavily on large-scale datasets~\cite{krizhevsky2012imagenet,xue2023alleviating,rajendran2024review}. However, well-trained models are known to memorize their training data~\cite{feldman2020does}, making them vulnerable to privacy attacks wherein adversaries can infer sensitive information by crafting sophisticated queries~\cite{fredrikson2015model,mia}. To address these risks, legislatures obligate the model owner to delete users' data upon receiving an unlearning request. Machine unlearning has emerged as a promising paradigm to remove specific data from a well-trained model, effectively erasing their influence as if they were never included in training~\cite{xu2023machine}. 

One challenge of machine unlearning is over-unlearning~\cite{over-unlearning}. The model's performance on the remaining data significantly deteriorates after unlearning. Several approaches have been proposed to address this issue. For example,~\cite{foster2024fast} suggest only modifying parameters associated with the unlearning samples to prevent affecting the remaining samples.~\cite{wang2022federated,tarun2023fast} fine-tune the unlearned model on the remaining data to recover the performance. However, existing methods require accessing the remaining samples, which renders them inapplicable in practice. We envision two critical requirements for a practical unlearning algorithm: \textbf{R1} \textit{It should work without accessing the remaining data (\textit{i.e.,} zero-shot)}; \textbf{R2} \textit{It should precisely remove only the targeted samples without leading to over-unlearning.} This task is challenging because knowledge is embedded in the model's weights and is highly interconnected~\cite{shwartz2017opening}. Adjustments to one part of the weights can inadvertently affect other parts. The zero-shot constraint further complicates the challenge.  

This paper proposes ZS-PAG (\underline{Z}ero-\underline{S}hot machine unlearning with \underline{P}roxy \underline{A}dversarial data \underline{G}eneration), a novel zero-shot machine unlearning method. In terms of \textbf{R1}, we perturb the unlearning samples over the decision boundaries and use the resulting samples to approximate the remaining samples. For \textbf{R2}, the unlearning should be more localized to the unlearning samples. Inspired by the phenomena that gradients lie in a subspace~\cite{li2018measuring}, we use the generated adversarial samples to approximate the inaccessible remaining samples and identify a corresponding subspace. The unlearning process is projected into the complementary subspace to prevent over-unlearning. In addition, we propose to optimize the pseudo-labels assigned to the unlearning samples using the influence function~\cite{koh2017understanding}. This ensures that unlearning the pseudo-labeled samples will have a positive influence on the remaining samples. In summary, our main contributions include:
\begin{itemize}
    \item \textit{Challenging problem}: We consider the zero-shot machine unlearning problem to represent the most practical and challenging setting for machine unlearning.
    
    \item \textit{Novel method}: We introduce a novel three-stage zero-shot machine unlearning framework, ZS-PAG, effectively preventing over-unlearning in a zero-shot context.

    \item \textit{Superior performance}: Experiments on various benchmarks demonstrate the efficacy of ZS-PAG. For instance, ZS-PAG outperforms the best baseline method by $6.03\%$ on the CIFAR-$100$ in a zero-shot setting. 
\end{itemize}

\begin{figure*}[tp]
    \centering
    \includegraphics[width=\linewidth]{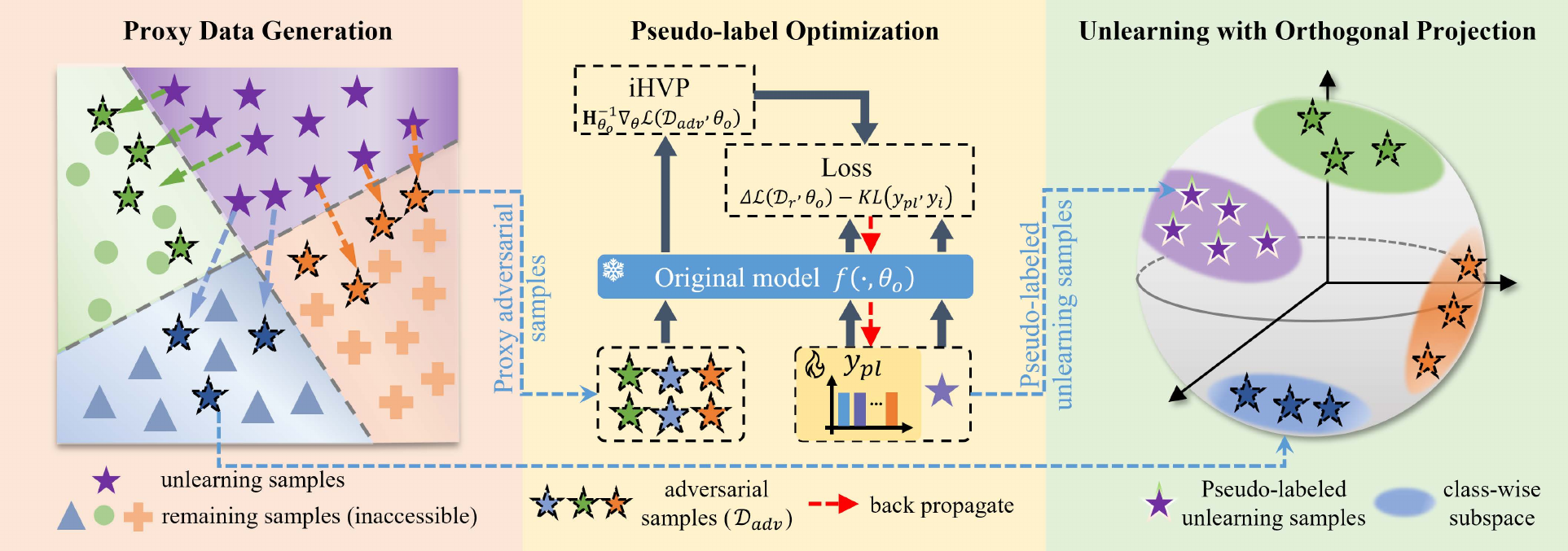}
    \caption{Framework of ZS-PAG. In the first step, we generate adversary samples to approximate the remaining samples. In the second step, we optimize the pseudo-labels assigned to unlearning samples to maximize the positive impact of the unlearning process on the remaining samples. In the third step, we project the unlearning process to a subspace orthogonal to the subspaces of the remaining classes.}
    \label{fig: pipline}
\end{figure*}

\section{Related Works}
\noindent\textbf{Machine unlearning}. Machine unlearning has received increasing attention over the past years~\cite{xu2023machine}. Current research can be broadly categorized into three areas: \textit{Sample-level unlearning}, \textit{feature-level unlearning}, and \textit{class-level unlearning}. \textit{Sample-level unlearning} focuses on the removal of specific training data points~\cite{bourtoule2021machine,cao2015towards,izzo2021appr,guo2020certified}. This is particularly relevant in scenarios where an individual requests the deletion of their data. \textit{Feature-level unlearning} targets the forgetting of particular features extracted from the training data~\cite{warnecke2021machine,li2023making}. This is useful when sensitive attributes like race or gender inadvertently influence the model. By unlearning these features, the model can mitigate biases while retaining its overall utility. \textit{Class-level unlearning} concentrates on making the model forget specific classes of data~\cite{chen2023boundary,tarun2023fast,golatkar2020eternal,fan2023salun,chang2024classmachineunlearningcomplex}. This is vital when ethical or organizational priorities demand eliminating outdated or harmful classifications. The objective is to remove all knowledge related to that class while maintaining the model's capability to handle other classes effectively.

\noindent\textbf{Mitigating over unlearning}. A significant challenge in machine unlearning is over-unlearning~\cite{over-unlearning,shen2024camu}, where the model's performance significantly degrades after unlearning. Various approaches have been proposed to address this issue.~\cite{chundawat2023can} employs a teacher-student framework, where the unlearned model is distilled from the original model. During distillation, only knowledge about the remaining samples is retained, while the information related to the unlearning samples is omitted.~\cite{bourtoule2021machine} proposes splitting the training data into multiple shards to isolate the influence of specific data points, ensuring their impact does not propagate to the remaining samples.~\cite{foster2024fast,fan2023salun} focus on selective model pruning to remove the targeted samples while preserving the model's utility. These methods identify critical parameters essential for general knowledge and selectively prune those linked to the targeted samples, ensuring the model's performance is preserved.~\cite{kurmanji2024towards} proposes fine-tuning the original model on both the remaining and unlearning datasets, using gradient negation for the latter to ensure effective unlearning without sacrificing performance.

However, existing methods assume access to the remaining data, which may be invalid. This paper differentiates itself from existing methods by considering unlearning in a zero-shot manner. While some existing works~\cite{chen2023boundary,chundawat2023zero,zhang2024efficientdatafreeunlearning} also explore zero-shot settings, they face the issue of over-unlearning, resulting in performance degradation on retained samples. In contrast, our method can accurately unlearn the targeted samples without causing over-unlearning. Our work shares a similar intent with~\cite{Hoang_2024_WACV,chen2024m,li2023sub}, but several key differences exist. Firstly, the research problems addressed are distinct. This paper focuses on zero-shot unlearning, which presents significant challenges due to the lack of remaining data. Secondly, alongside addressing the issue of over-unlearning, we investigate how unlearning can potentially enhance the model's performance by optimizing the influence.

\section{Proposed Method}
\subsection{Notations}
Let $f: \mathbb{R}^d\mapsto \mathbb{R}^C$ be a classifier parameterized by $\theta\coloneqq \{\mathbf{w}^l\}_{l=1}^L$, and $\mathbf{r}_l^i$ is the input feature w.r.t input $x_i$ at the $l$-th layer. Define $\theta_o$ as an empirical minimizer by fitting $f$ on the training set $\mathcal{D} = \left\{z_i\coloneqq(x_i,y_i)\right\}_{i=1}^N$ with some loss function $\ell(x,y;\theta)$. The training data $\mathcal{D} = \mathcal{D}_u \cup \mathcal{D}_r$ can be partitioned into two disjoint sets of unlearning data $\mathcal{D}_u$ and remaining data $\mathcal{D}_r$. Existing methods assume access to both $\mathcal{D}_r$ and $\mathcal{D}_u$. However, this assumption may be invalid in practice. In this paper, we consider a more stringent setting where we only have access to $f(\theta_o)$ and $\mathcal{D}_u$. Figure \ref{fig: pipline} provides an overview of the proposed framework.

\subsection{Proxy Adversarial Data Generation} \label{sec:gene_adv}
In zero-shot machine unlearning, we only have unlearning samples. However, it is hard to prevent over-unlearning without knowledge about the remaining samples. As such, our first step is to generate adversary samples as a proxy for the inaccessible remaining samples. 

Starting with an unlearning sample $x_i\in \mathcal{D}_u$, we generate an adversary sample $x_{adv}=x_i+\delta$ via optimizing the additive noise $\delta$. The goal is to deceive the original model into predicting $x_{adv}$ as a pre-specified class $y_{target} \in \Bar{c}\coloneqq [C]\backslash y_i$ in a $C$ classification task. The optimization problem is formulated as follows:
\begin{equation} \label{eq:tat}
    \min_{\delta}\mathcal{L}(x_i+\delta,y_{target};\theta), \quad \text{s.t.} \quad \|\delta\|_p\le \epsilon.
\end{equation}
The target label is determined by:
\begin{equation} 
    y_{target} = \arg\max_{k\in \Bar{c}}f_k(x_i;\theta_o),
\end{equation}
where $f_k(x_i;\theta_o)$ is the $k$-th component of logits. We use the second high prediction $k$ as the $y_{target}$ for the adversary sample. This choice prioritizes efficiency as perturbing the sample towards this class typically requires less modification compared to forcing a more distant prediction. Essentially, we are perturbing $x_i$ cross decision boundaries. Repeating this process for each unlearning sample, we end with a set of adversary samples residing in regions of the rest classes, which serve as the basis for the subspace estimation in Section \ref{sec:fss}. Directly solving Eq.~\ref{eq:tat} is prohibitively hard due to high non-convexity. A plethora of algorithms have been proposed to approximate the solution~\cite{zhou2022adversarial}. We use the method proposed by~\cite{madry2017towards} in this paper.

\subsection{Unlearning with Orthogonal Projection} \label{sec:fss}
Existing research reveals that gradients lie in a lower-dimensional subspace spanned by the inputs~\cite{li2022low,GPM}. Inspired by this, we project the unlearning gradients onto a designated subspace where the gradients are irrelevant to the remaining classes. This design effectively prevents over-unlearning. 

We identify the subspace of class $k$ by feeding a batch of adversary samples $\mathcal{B}_k=\{(x_i, y_i)\mid (x_i, y_i) \in \mathcal{D}_{adv}, y_i = k \}_{i=1}^{n_{adv}}$ through the original model $f(\theta_o)$. It is expected that using different values of $n_{adv}$ will produce various subspaces and affect the performance of the proposed method. However, as demonstrated in the experiment, setting $n_{adv}=100$ is sufficient for our needs. At each layer, we compute the subspace corresponding to class $k$ at layer $l$ as follows: 
\begin{equation} \label{eq:liner_svd}
    \mathbf{U}_l^k\mathbf{\Sigma}_l^k\mathbf{V}_l^{kT}=\text{SVD}(\mathbf{R}_l^k),
\end{equation}
where $\mathbf{U}_l^k$ contains the basis of the subspace,  $\mathbf{R}_l^k=[\mathbf{r}_{l,1}, \mathbf{r}_{l,2},..., \mathbf{r}_{l,n_{adv}}]$ represents the input feature maps. The definition of $\mathbf{R}_l^k$ varies by layer type. For fully connected layers, $\mathbf{R}_l^k$ is the raw input feature map. In convolutional layers, $\mathbf{R}_l^k$ is a reshaped input feature map where each row concatenates multiple convolutional patches of the raw input~\cite{liu2018frequency}. For attention layers consisting of three modules, \textit{i.e.,} ($\mathbf{W}_q$, $\mathbf{W}_k$, and $\mathbf{W}_v$), we use the raw input of each module as $\mathbf{R}_l^k$ and compute a separate subspace for each module. 

To prevent over-unlearning, we project the unlearning gradients onto a subspace orthogonal to the subspaces of the remaining classes. Considering unlearn class $c$, we iterate over the rest classes $k \in \Bar{c}$ to calculate the combined subspace $\mathcal{S}^{\Bar{c}}$ as: 
\begin{equation} \label{eq: proj mat}
    \mathcal{S}^{\Bar{c}}_l = \text{SVD}(\texttt{Concatenate}(\mathcal{S}^{\Bar{c}}_l,\mathbf{U}_l^k)),
\end{equation}
where $\mathcal{S}^{\Bar{c}}$ is initialized as an empty set and $\mathcal{S}^{\Bar{c}}_l$ is the $l$-th element. The projection matrix is defined as $\mathbf{P}_l = \mathbf{I}-\mathcal{S}_l^{\Bar{c}}\mathcal{S}_l^{\Bar{c}T}$.
At layer-$l$, the updating rule is:
\begin{equation} \label{eq: psgd}
    \mathbf{w}_{t+1}^l=\mathbf{w}_{t}^l-\eta \mathbf{P}_l^{\Bar{c}} \nabla_{\mathbf{w}^l}\ell.
\end{equation}
The following theorem shows unlearning with Eq. \ref{eq: psgd} prevents over-unlearning.


\begin{restatable}{theorem}{nullspace}

    Denote $f$ a neural network parameterized by $\theta\coloneqq \{\mathbf{w}^l\}_{l=1}^L$, where each $\mathbf{w}^l$ represents the weights of the $l$-th layer. Assume the loss function $\mathcal{L}_{r}(\theta) \coloneqq \frac{1}{|\mathcal{D}_r|}\sum_{\mathcal{D}_r}\ell(f(x_i;\theta),y_i)$ has a Lipschitz continuous gradient with constant $L$ and satisfies the Polyak-Łojasiewicz (PL) inequality~\cite{POLYAK1963864} with a constant $\mu$. Under suitable conditions on the learning rate, unlearning with Eq.~\ref{eq: psgd} ensures the unlearned model $f(\theta_u)$ will have approximately the same performance as $f(\theta_o)$, \textit{i.e.},
    \begin{equation}
        \mathcal{L}_{r}(\theta_u) \approx \mathcal{L}_{r}(\theta_o).
    \end{equation}
\end{restatable}

\subsection{Influence-based Pseudo-label Optimization} \label{sec:opt if}
In addition to preventing over-unlearning, we also propose optimizing the unlearning process's influence on the remaining samples $\mathcal{D}_r$. Denote $\theta^* = \arg\min_{\theta}\frac{1}{N}\sum_{n=1}^{N}\ell(z_n; \theta)$ as the empirical risk minimizer. After removing a point $z_i$ from the training set and the optimal parameters became  $\theta^*_{-z_i} = \arg\min_{\theta}\frac{1}{N}\sum_{z_n\neq z_i}\ell(z_n; \theta)$. The influence function gives an efficient approximation to the weight change $\theta^*-\theta^*_{-z_i}$. 
\begin{definition}[Influence function]~\cite{koh2017understanding} When upweighting $z_i\coloneqq(x_i,y_i)$ by $\varepsilon$, we get new parameter as $\theta^*_{\varepsilon,z_i}=\arg\min_{\theta}\frac{1}{N}\sum_{n=1}^{N}\ell(z_n; \theta)+\epsilon\ell(z_i;\theta)$. The influence of upweighting $z_i$ on the parameters $\theta^*$ is given by
\begin{equation} \label{eq: if}
    \mathcal{I}_{z_i}=\left.\frac{d\theta^*_{\varepsilon,z_i}}{d\varepsilon}\right|_{\varepsilon=0} = -\mathbf{H}_{\theta^*}^{-1}\nabla_{\theta}\ell(z_i;\theta^*),
\end{equation}
where $\mathbf{H}_{\theta^*}=\frac{1}{N}\sum_{n=1}^{N}\nabla_{\theta}^2\ell(z_n; \theta^*)$ is the Hessian of the loss function on the training data. 
\end{definition}
Assume we have access to $\mathcal{D}_r$ for now. Consider upweighting a pseudo-labeled unlearning sample $z_i\coloneqq (x_i, y_{pl})$ by $\varepsilon$. The change in loss of the remaining sample $x \in \mathcal{D}_r$ is estimated as:
\begin{equation}
\begin{split}
   \Delta \mathcal{L}(\mathcal{D}_r, \theta_u)&=\mathcal{L}(\mathcal{D}_r, \theta_u) - \mathcal{L}(\mathcal{D}_r, \theta_o) \\& \approx \nabla_\theta \mathcal{L}(\mathcal{D}_r, \theta_o)^T (\theta_u-\theta_o)\\
   & \approx \nabla_\theta \mathcal{L}(\mathcal{D}_r, \theta_o)^T \varepsilon\mathcal{I}_{z_i},
\end{split}
\end{equation}
where the last step is the application of influence function $\mathcal{I}_{z_i}$. Replace $\mathcal{I}_{x_i}$ with Eq. \ref{eq: if}, we get:
\begin{equation}
      \Delta \mathcal{L}(\mathcal{D}_r, \theta_u)\approx -\varepsilon\nabla_\theta \mathcal{L}(\mathcal{D}_r, \theta_o)^T \mathbf{H}_{\theta_o}^{-1}\nabla_{\theta}\ell(x_i,y_{pl};\theta_o).
\end{equation}
Removing a sample $(x_i,y_{pl})$ is equivalent to setting $\varepsilon=-\frac{1}{N}$. Therefore, we approximate the change of loss on $\mathcal{D}_r$ after unlearning $x_i$ as:
\begin{equation}\label{eq: inf_loss}
    \Delta \mathcal{L}(\mathcal{D}_r, \theta_u) \approx \frac{1}{N}\nabla_\theta \mathcal{L}(\mathcal{D}_r, \theta_o)^T \mathbf{H}_{\theta_o}^{-1}\nabla_{\theta}\ell(x_i, y_{pl};\theta_o).
\end{equation}
Calculating $\mathbf{H}_{\theta_o}^{-1}$ is costly for DNNs. Therefore, iterative methods~\cite{sattigeri2022fair} are often used in practice to approximate $\mathbf{H}_{\theta_o}^{-1}\nabla_\theta \mathcal{L}(\mathcal{D}_r, \theta_o)$. 


We random initialize $y_{pl} \in \mathbb{R}^C$ and optimize it to decrease $\Delta \mathcal{L}(\mathcal{D}_r, \theta_u)$ approximated by Eq. \ref{eq: inf_loss}. However, minimizing $\Delta \mathcal{L}(\mathcal{D}_r, \theta_u)$ alone may lead $y_{pl}$ collapse into the ground-truth label $y_i$.  As such, we add a regularization term to penalize the similarity between $y_{pl}$ and $y_i$. This could be done via any similarity metric. Our experiment empirically finds that the Kullback-Leibler (KL) divergence suffices our needs. By optimizing $y_{pl}$, we ensure the model will benefit from the unlearning process. This results in improved performance after unlearning. In a zero-shot unlearning setting, we approximate $\mathcal{D}_r$ with $\mathcal{D}_{adv}$ generated as described in Section \ref{sec:gene_adv}. Our empirical results demonstrate that $\mathcal{D}_{adv}$ serves as an adequate proxy for $\mathcal{D}_r$.

\begin{table*}[htp]
    \centering
    \setlength{\tabcolsep}{3pt}
    \begin{tabular}{c|cc|cc|cc|cc}
    \toprule
   \multirow{2}{*}{Approach} & \multicolumn{2}{c|}{Facescrub/ResNet} & \multicolumn{2}{c|}{SVHN/VGG} & \multicolumn{2}{c|}{CIFAR-10/ViT} & \multicolumn{2}{c}{CIFAR-100/ResNet}\\
    & $Acc_{\mathcal{D}_{rt}}(\uparrow)$ & $Acc_{\mathcal{D}_{ut}}(\downarrow)$ &   $Acc_{\mathcal{D}_{rt}}(\uparrow)$ & $Acc_{\mathcal{D}_{ut}}(\downarrow)$ &$Acc_{\mathcal{D}_{rt}}(\uparrow)$ & $Acc_{\mathcal{D}_{ut}}(\downarrow)$&
    $Acc_{\mathcal{D}_{rt}}(\uparrow)$ & $Acc_{\mathcal{D}_{ut}}(\downarrow)$\\
    \midrule 
    Original & $96.17_{\pm 0.05}$ & $96.05_{\pm 3.20}$  & $95.52_{\pm0.12}$ & $91.30_{\pm0.30}$ & $ 83.80_{\pm 1.16} $ & $63.97_{\pm 0.46}$ &  $73.31_{\pm 0.31}$ & $72.07 _{\pm 1.18}$  \\

    Retrain & $96.31_{\pm 0.15}$ & $0.00 _{\pm0.00}$ & $95.56_{\pm0.23}$ & $0.00 _{\pm0.00}$ & $86.57_{\pm0.28} $ & $0.00_{\pm 0.00} $ & $75.36_{\pm 0.34} $ & $0.00_{\pm0.00}$ \\
    \midrule
    
    FT &$74.69_{\pm 6.31}$ & $0.00_{\pm 0.00}$ &  $95.97_{\pm 0.05}$ & $0.95_{\pm 0.51}$ & $87.60_{\pm 0.65}$ & $6.43_{\pm 1.68}$ & $74.68_{\pm 0.03}$ & $2.33_{\pm 0.94}$ \\
    
    \rowcolor{lightgray}
    FT-zs &$78.72_{\pm 0.25}$ & $0.00_{\pm 3.34}$ & $80.41_{\pm 6.60}$ & $5.91_{\pm 2.80}$ & $80.69_{\pm 1.64}$ & $4.70_{\pm 1.55}$ & $ 66.63_{\pm 0.22}$ & $0.33_{\pm 0.47}$ \\

    Neggrad &$65.59_{\pm 2.10}$ & $0.00_{\pm 0.00}$ & $96.15_{\pm 0.08}$ & $0.02_{\pm 0.03}$ & $88.29_{\pm 0.63}$ & $4.10_{\pm 2.05}$ & $ 75.15_{\pm 0.15}$ & $3.00_{\pm 1.63}$  \\
    
    \rowcolor{lightgray}
    Neggrad-zs &$79.09_{\pm 1.05}$ & $3.85_{\pm 0.33}$ & $57.02_{\pm 34.14}$ & $9.40_{\pm 13.30}$  & $76.89_{\pm 7.71}$ & $1.77_{\pm 1.53}$  & $66.01_{\pm 4.88}$ & $2.33_{\pm 3.30}$ \\
    
    BadT &$87.51_{\pm 1.03}$ & $6.41_{\pm 4.80}$& $96.01_{\pm 0.06}$ & $3.27_{\pm 0.65}$  & $85.51_{\pm 1.29}$ & $2.20_{\pm 1.82}$& $71.94_{\pm 0.32}$ & $8.33_{\pm 2.49}$\\

    \rowcolor{lightgray}
    BadT-zs &$74.60_{\pm 1.60}$ & $5.77_{\pm 2.72}$ & $78.22_{\pm 4.63}$ & $2.30_{\pm 0.67}$ & $68.54_{\pm 4.58} $ & $4.23_{\pm 2.03}$  & $24.69_{\pm 16.71}$ & $8.67_{\pm 6.34}$\\
 
    SalUn &$85.20_{\pm 1.63}$ & $1.28_{\pm 1.81}$ & $95.85_{\pm 0.06}$ & $3.48_{\pm 1.14}$ & $87.51_{\pm 0.75}$ & $6.47_{\pm 1.62}$ & $74.68_{\pm 0.10}$ & $1.67_{\pm 1.25}$ \\
    
    \rowcolor{lightgray}
    SalUn-zs &$77.50_{\pm 0.63}$ & $0.00_{\pm 0.00}$ & $80.38_{\pm 6.62}$ & $5.86_{\pm 2.66}$ & $80.77_{\pm 1.81}$ & $4.60_{\pm 1.39}$ & $67.82_{\pm 1.41}$ & $0.00_{\pm 0.00}$ \\

    Fisher &$51.88_{\pm 1.36}$ & $0.00_{\pm 0.00}$ & $92.69_{\pm 1.03}$ & $0.54_{\pm 0.67}$ & $84.51_{\pm 1.35}$ & $6.27_{\pm 7.1}$ & $50.03_{\pm 1.93}$ & $0.00_{\pm 0.00}$  \\

    \rowcolor{lightgray}
    BU &$77.21_{\pm 1.98}$ & $0.64_{\pm 0.91}$ & $82.19_{\pm 6.52}$ & $9.26_{\pm 4.06}$ & $80.78_{\pm 1.72}$ & $4.67_{\pm 1.55}$ & $69.51_{\pm 0.39}$ & $3.00_{\pm 0.82}$ \\

    \rowcolor{lightgray}
    GKT & $73.10_{\pm 2.52}$ & $0.99_{\pm 0.17}$ & $ 94.25_{\pm 0.79}$ & $2.95_{\pm 2.14} $ & $ 74.67_{\pm 6.35}$ & $9.83_{\pm 5.97}$ & $ 63.53_{\pm 0.78}$ & $6.00_{\pm 6.48}$ \\
 
    \midrule
    \rowcolor{lightgray}
    ZS-PAG &$96.48_{\pm 0.22}$ & $1.92_{\pm 0.21}$ & $95.63_{\pm 0.07}$ & $0.21_{\pm 0.11}$  & $85.47_{\pm 1.11}$ & $1.40_{\pm 0.14}$ &  $75.54_{\pm 0.33}$ & $2.00_{\pm 1.63}$\\
    \bottomrule
    \end{tabular} %
    \caption{\textbf{Single-class unlearning.} Comparison of ZS-PAG and baselines across various datasets and model architectures. Results highlighted in light gray are obtained in a zero-shot setting.}
    \label{tab: main_res}
\end{table*}

\section{Experiment} \label{sec:exp}
\subsection{Experimental Setup}
\noindent{\textbf{Dataset and model architecture.}} Following previous works, we evaluate the proposed method on four benchmarks: Facescrub~\cite{ng2014data}, SVHN~\cite{netzer2011reading}, CIFAR-$10$ and CIFAR-$100$~\cite{krizhevsky2009learning}. We apply four representative network architectures in our experiments: AlexNet~\cite{krizhevsky2012imagenet}, VGG~\cite{simonyan2014very}, ResNet~\cite{he2015deep}, and ViT~\cite{dosovitskiy2020image}. 

\noindent{\textbf{Baselines.}} We compare our approach with several baselines. The first five methods assume access to $\mathcal{D}_r$: (1) FT~\cite{warnecke2021machine}, (2) Neggrad~\cite{kurmanji2024towards}, (3) BadT~\cite{chundawat2023can}, (4) SalUn~\cite{fan2023salun}, and (5) Fisher~\cite{golatkar2020eternal}. To make a fair comparison, we included the results of their zero-shot versions. We also add two baselines that do not use $\mathcal{D}_r$: (6) BU~\cite{chen2023boundary} and (7) GKT~\cite{chundawat2023zero}. Finally, we include the results of (8) Retrain. 

\noindent{\textbf{Evaluation metrics and implement details.}} Following the literature, we assess the unlearned model with three metrics: 1) $\text{Acc}_{ut}$: Accuracy on the testing set of unlearning classes. In a class unlearning setting, the unlearned model should have zero accuracy on the unlearning classes, matching a retrained model; 2) $\text{Acc}_{mia}$: Accuracy of membership inference attack (MIA). We train an attack model to predict the membership of unlearning samples in the training set. As noted by~\cite{fan2023salun,chen2023boundary}, the closer this metric is to that of the retrained model, the better the performance of the unlearning algorithm; 3) $\text{Acc}_{rt}$: Accuracy on testing set of remaining classes. We measure the degree of over-unlearning. An idea unlearned model should not decrease $\text{Acc}_{rt}$. All results are averaged over three different runs. We utilize projected gradient descent~\cite{madry2017towards} to generate adversary samples $\mathcal{D}_{adv}$ in the experiment. Note our method is compatible with any adversary attack strategy.
 
\subsection{Single-class Unlearning}\label{sec: sc_unlearn} We randomly select one class for SVHN and CIFAR10, two classes for Facescrub, and ten classes for CIFAR100 as the unlearning class(es) and compare our method with the baselines. Results in Table \ref{tab: main_res} reveal the following key findings:

1) \textit{Existing unlearning methods rely on $\mathcal{D}_r$ to prevent over-unlearning.} There exists a significant gap in $Acc_{\mathcal{D}_{ut}}$ between these methods and their zero-shot versions. For instance, on the SVHN dataset, the FT method fine-tunes the original model using a combination of $\mathcal{D}_r$ and random labeled samples from $\mathcal{D}_u$. As a result, $Acc_{\mathcal{D}_{ut}}$ of the unlearned model is decreased to $0.95\%$, while $Acc_{\mathcal{D}_{rt}}$ remains at $95.97\%$, which is comparable to the original model. However the $Acc_{\mathcal{D}_{rt}}$ of FT-zs drops by more than $15.5\%$. This comparison highlights FT's reliance on $\mathcal{D}_r$ to prevent over-unlearning. This conclusion also applies to Neggrad, BadT, and SalUN. 

2) \textit{ZS-PAG out-stands in zero-shot unlearning}. For example, on the CIFAR-$10$ dataset, $Acc_{\mathcal{D}_{ut}}$ and $Acc_{\mathcal{D}_{rt}}$ of ZS-PAG are $1.40\%$ and $85.47\%$, respectively. In comparison, the original model yields $63.97\%$ and $83.80\%$, respectively. ZS-PAG effectively removed the information of the unlearning classes from the model and improved the model's performance in the remaining class. The improvement in $Acc_{\mathcal{D}_{rt}}$ can be attributed to the optimized pseudo-labels in ZS-PAG, which maximize the influence of the unlearning process on the remaining samples. Notably, ZS-PAG improves $Acc_{\mathcal{D}_{rt}}$ by $1.53\%$, $1.67\%$, and $2.23\%$ on the FashionMNIST, CIFAR-$10$, and CIFAR-$100$ dataset, respectively.

\begin{table}[ht]
    \centering

    \begin{tabular}{c|ccc}
    \toprule
        Approach & $Acc_{\mathcal{D}_{rt}}(\uparrow)$ & $Acc_{\mathcal{D}_{ut}}(\downarrow)$ & $Acc_{mia}$  \\
        \midrule
    
    Original & $95.39_{\pm 0.15}$ & $94.41 _{\pm 0.13}$ & $67.57_{\pm 0.55}$ \\

    Retrain & $95.90_{\pm 1.15}$ & $0.00_{\pm 0.00}$ & $39.38_{\pm 4.86}$\\
    \midrule
 
    FT & $96.67_{\pm 0.07}$ & $0.03_{\pm 0.01}$ & $36.28_{\pm 4.15}$ \\

    \rowcolor{lightgray}
    FT-zs & $52.38_{\pm 6.13}$ & $0.23_{\pm 0.12}$ & $51.30_{\pm 2.61}$  \\
    
    Neggrad &  $95.06_{\pm 2.74}$ & $0.00_{\pm 0.00}$ & $52.44_{\pm 0.90}$  \\

    \rowcolor{lightgray}
    Neggrad-zs & $32.43_{\pm 16.99}$ & $1.24_{\pm 1.75}$ & $49.73_{\pm 1.11}$ \\

    BadT  & $92.96_{\pm 0.55}$ & $0.61_{\pm 0.87}$ & $64.00_{\pm 5.13}$  \\

    \rowcolor{lightgray}
    BadT-zs  & $18.48_{\pm 4.61}$ & $0.25_{\pm 0.35}$ & $40.61_{\pm 3.78}$ \\

    SalUn  & $96.48_{\pm 0.12}$ & $0.14_{\pm 0.02}$ & $40.82_{\pm 4.20}$ \\

    \rowcolor{lightgray}
    SalUn-zs  & $52.48_{\pm 6.14}$ & $0.24_{\pm 0.14}$ & $49.91_{\pm 2.24}$ \\

    Fisher & $95.18_{\pm 0.07}$ & $12.36_{\pm 10.91}$ & $50.06_{\pm 8.57}$ \\

    \rowcolor{lightgray}
    BU & $53.64_{\pm 5.75}$ & $0.45_{\pm 0.23}$ & $47.95_{\pm 0.90}$ \\
    
    \rowcolor{lightgray}
    GKT  & $  72.78_{\pm 3.14} $ & $0.00_{\pm 0.00}$ & $43.19_{\pm 0.76} $ \\
    \midrule
    
    \rowcolor{lightgray}
    ZS-PAG & $96.58_{\pm 0.18}$ & $0.31_{\pm 0.14}$ & $47.81_{\pm 3.60}$\\
    \bottomrule
    \end{tabular}
    \caption{\textbf{Multi-class unlearning}. We randomly unlearn $3$ classes from  SVHN and compare our method against baselines. Results highlighted in light gray are obtained in a zero-shot setting.}
    \label{tab: mc_unlearn}
\end{table}

\subsection{Multi-class Unlearning} We extend the comparison to multi-classes unlearning by randomly unlearning $3$ classes from the SVHN dataset and comparing ZS-PAG against baselines. As shown in Table \ref{tab: mc_unlearn}, the baseline methods exhibit a similar performance drop in $Acc_{\mathcal{D}{rt}}$ as seen in Table \ref{tab: main_res}. What's worse, the over-unlearning issue worsens when unlearning more classes. For instance, the gap of $Acc_{\mathcal{D}_{rt}}$ between BadT and BadT-zs is $ 17.79\%$ in Table ~\ref{tab: main_res} for single class unlearning. And this gap increases to $74.48\%$ when unlearning $3$ classes. In contrast, ZS-PAG successfully prevents over-unlearning and demonstrates comparable even higher $Acc_{\mathcal{D}_{rt}}$ than baselines, despite these methods using remaining samples during unlearning.

Comparing ZS-PAG to the original model across Tables \ref{tab: main_res} and \ref{tab: mc_unlearn}, we observe increasing gains in $Acc_{\mathcal{D}_{rt}}$ as more classes are unlearned: $0.11\%$ for $1$ class and $1.19\%$ for $3$classes. We accumulate a greater impact from each unlearning sample on the remaining samples when unlearning more classes, resulting in an improved overall outcome.

\begin{figure*}[htp]
    \centering
    \begin{subfigure}[b]{0.23\textwidth}
        \includegraphics[width=\textwidth]{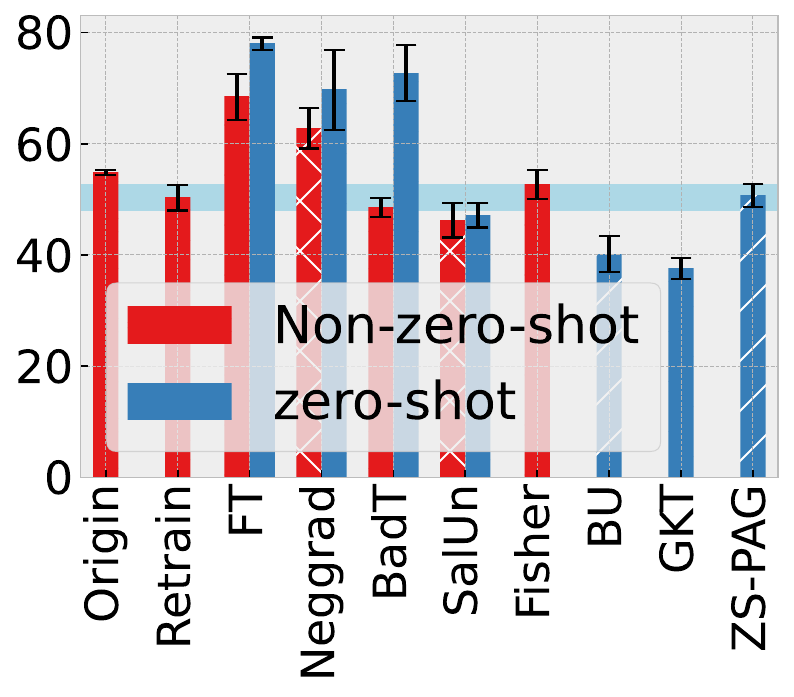}
        \caption{Facescrub}
    \end{subfigure}
    \hfill
    \begin{subfigure}[b]{0.23\textwidth}
        \includegraphics[width=\textwidth]{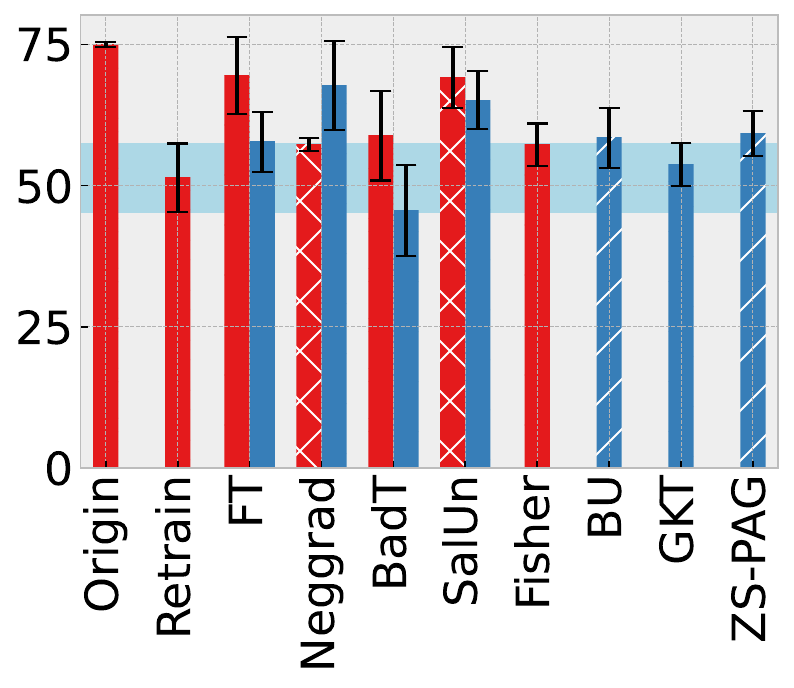}
        \caption{SVHN}
        \label{fig: svhn mia}
    \end{subfigure}
    \hfill
    \begin{subfigure}[b]{0.23\textwidth}
        \includegraphics[width=\textwidth]{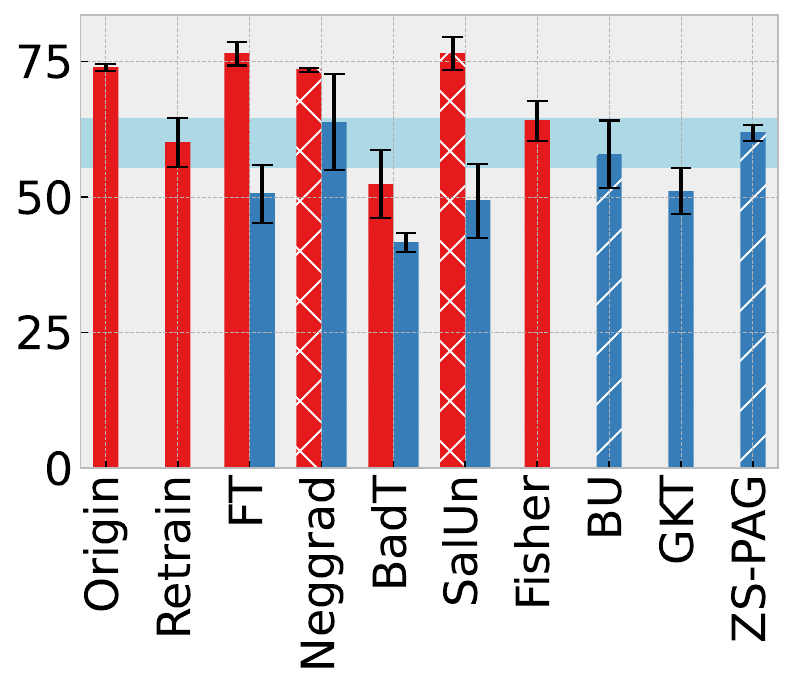}
        \caption{CIFAR-$10$}
    \end{subfigure}
    \hfill
    \begin{subfigure}[b]{0.23\textwidth}
        \includegraphics[width=\textwidth]{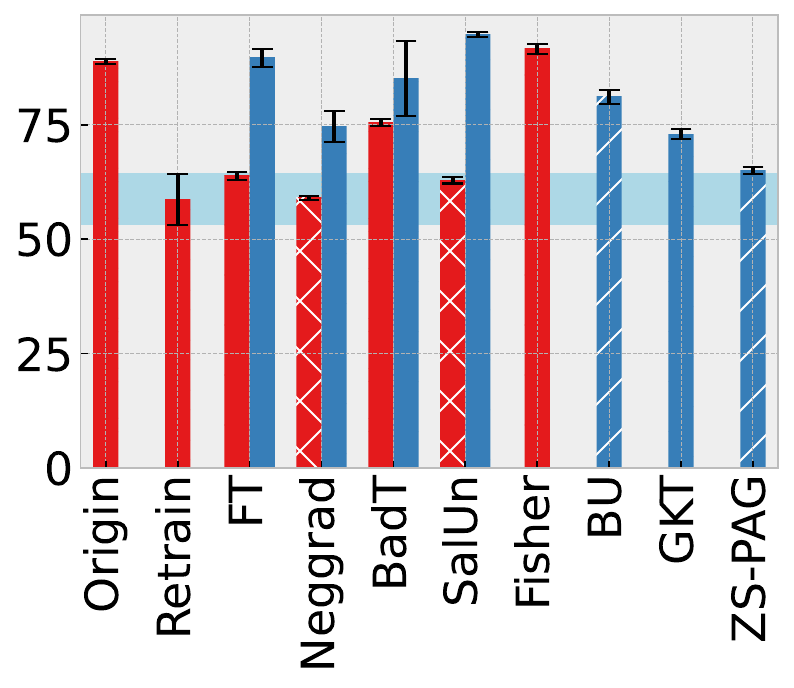}
        \caption{CIFAR-$100$}
    \end{subfigure}
    \caption{MIA results of ZS-PAG and baselines for single-class unlearning. Results outside the highlighted optimal region may leak information about the unlearning samples.}
    \label{fig: mia}
\end{figure*}

\begin{figure}[htp]
    \centering
    \includegraphics[width=\linewidth]{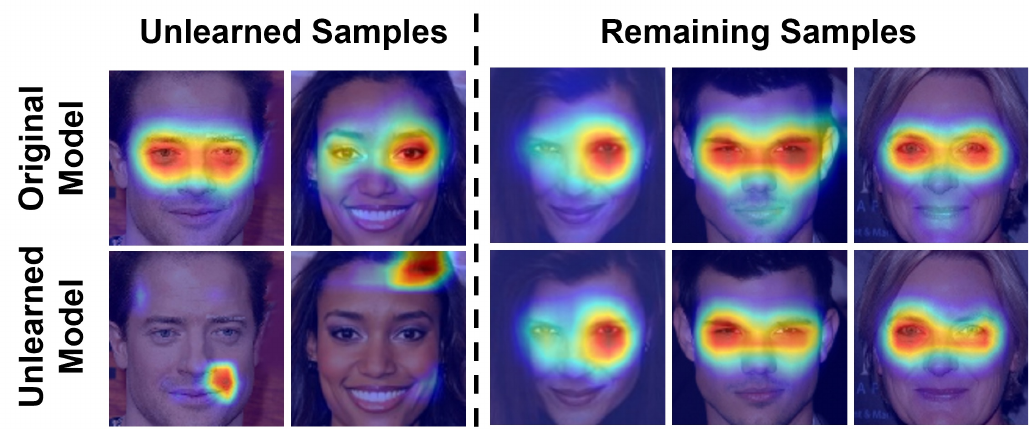}
    \caption{GradCAM results on unlearned and remaining samples for the original and unlearned models obtained by ZS-PAG.}
    \label{fig: gradcam}
\end{figure}

\subsection{Unlearning Guarantee}
\noindent{\textbf{MIA}.} Following the prior arts~\cite{fan2023salun,chen2023boundary}, we use the same setting as Sec. \ref{sec: sc_unlearn} and perform MIA against the unlearned model to probe the retained information about the unlearning classes in the unlearned model. Results are presented in Figure \ref{fig: mia}. Note a large deviation of $Acc_{mia}$ from the retrained model may leak information about the unlearning samples. Therefore, $Acc_{mia}$ values closer to the retrained model are more desirable. The optimal regions are highlighted in Figure \ref{fig: mia}. The baselines exhibit $Acc_{mia}$ values either significantly higher or lower than that of the retrained model. For example, on the SVHN dataset in Figure \ref{fig: svhn mia}, FT achieves the highest $Acc_{mia}$ among all unlearning methods. This indicates an unsuccessful unlearning. In contrast, the results of ZS-PAG lie close to the optimal region on all datasets, suggesting a successful unlearning.

\noindent{\textbf{Visualize attention map}.} We plot the GradCAM~\cite{selvaraju2017grad} results on samples from the Facescrub dataset to evaluate the effectiveness of ZS-PAG. Two classes were unlearned from the original model to obtain the unlearned model. The results in Figure ~\ref{fig: gradcam} demonstrate the effectiveness of our unlearning approach in selectively erasing target samples while preserving the model's performance on the remaining data. The first row shows heatmaps for the original model, and the second row corresponds to the unlearned model. For unlearning samples, the original model focuses on the eye regions, which are critical for identity recognition, whereas the unlearned model displays random attention, indicating successful forgetting. For the remaining samples, both models exhibit consistent attention to the eye regions, demonstrating that ZS-PAG preserves the model’s functionality on non-target data.

\noindent{\textbf{Backdoor attack-based metric.}} Beyond MIA and attention maps, we additionally employ a backdoor attack-based metric to further evaluate the unlearning effectiveness of ZS-PAG and its robustness against over-unlearning, following the protocol in \cite{chundawat2023zero}. Specifically, we train a ViT model on the CIFAR-10 dataset, implant a backdoor by adding visual triggers to samples from class 1, and relabel them as class 8 to simulate a targeted attack scenario. We apply ZS-PAG to unlearn class 1 and remove its influence. The results, shown in Tab.~\ref{tab:backdoor}, demonstrate that the attack accuracy ($Acc_{attack}$) drops dramatically from 89.09 to 0.24 after unlearning. This significant reduction indicates the successful elimination of the backdoor functionality and further confirms the effectiveness of ZS-PAG in erasing sensitive information while preserving the integrity of the remaining model.

\begin{table}[htp]
    \centering
    \begin{tabular}{c|ccc}
        \toprule
         CIFAR10/ViT  & Original & Retrain & ZS-PAG  \\
         \midrule
         $Acc_{D_{rt}}(\uparrow)$ & $84.12$ & $83.96$ & $85.33$  \\
         $Acc_{D_{ut}}(\downarrow)$ & $67.20$ & $0.00$ & $0.96$ \\
         $Acc_{attack}(\downarrow)$ & $89.09$ & $-$ & $0.24$ \\
         \bottomrule 
    \end{tabular}
    \caption{\textbf{Backdoor attack-based metric.} We unlearn $1$ class from a backdoored ViT.}
    \label{tab:backdoor}
\end{table}

\subsection{Ablation Studies}
In this section, we analyze key components of the proposed ZS-PAG framework. These studies focus on (1) the distribution of adversarial samples, (2) the influence of adversarial attack, and (3) the contributions of subspace projection and pseudo-labeling. The results provide insights into the robustness and effectiveness of the proposed method.
\begin{figure}[t]
    \centering
    \begin{subfigure}[b]{0.23\textwidth}
        \includegraphics[width=1\linewidth]{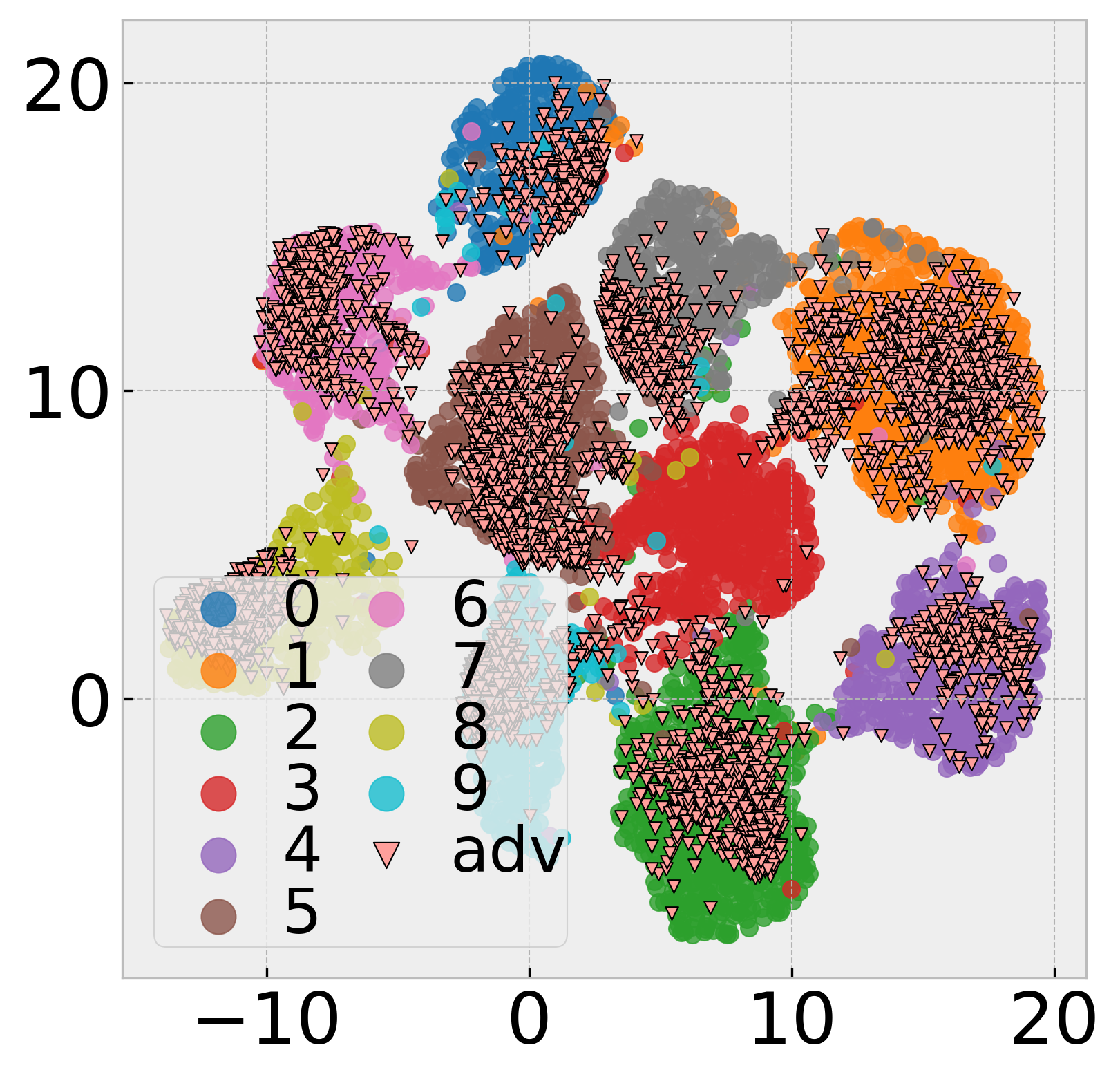}
        \caption{SVHN}
    \end{subfigure}
    \hfill
    \begin{subfigure}[b]{0.23\textwidth}
        \includegraphics[width=1\linewidth]{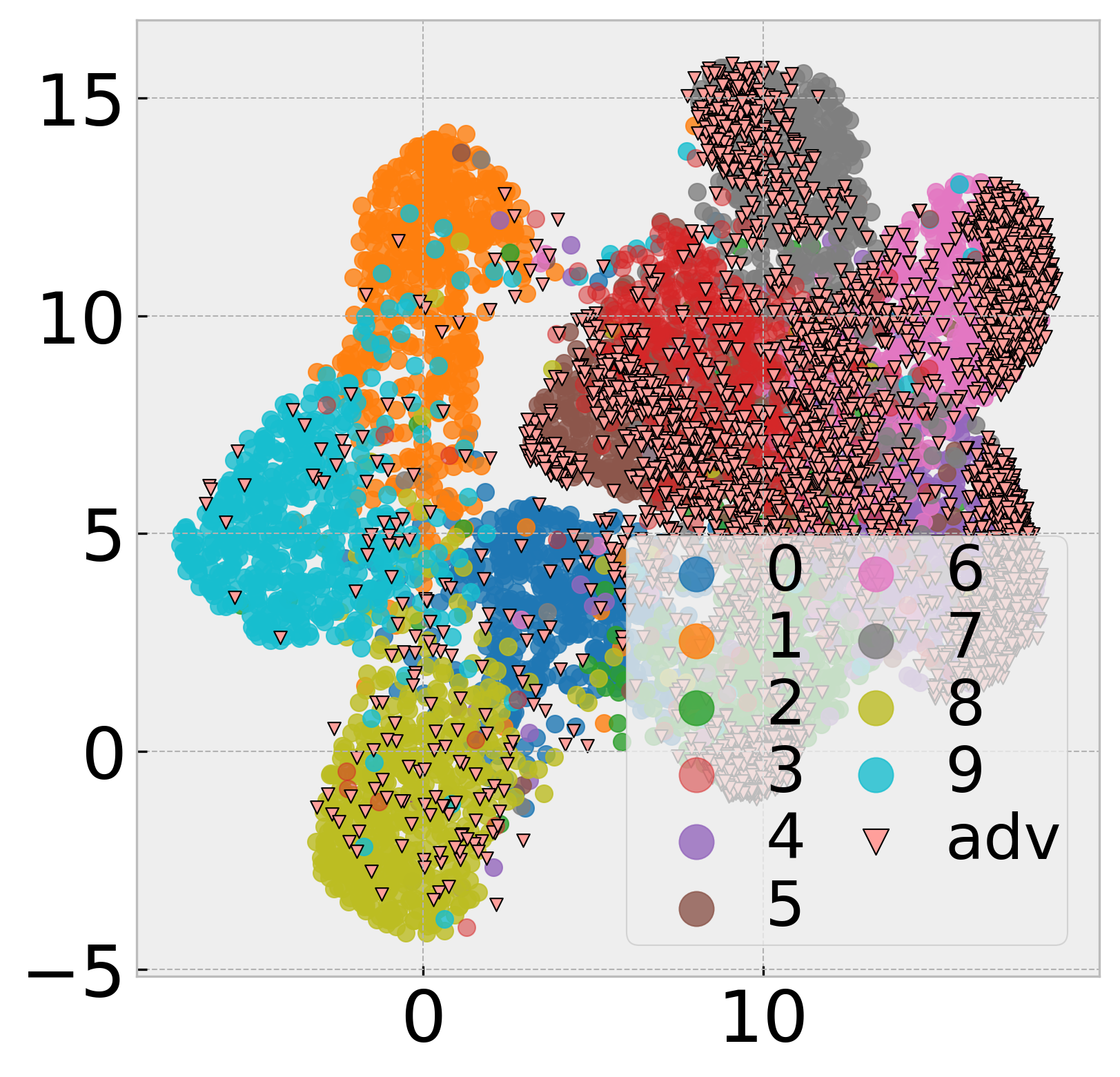}
        \caption{CIFAR-$10$}
    \end{subfigure}
    \caption{Distribution of generated adversarial samples. We set class $3$ as the unlearning class and generate adversarial samples to approximate samples of the rest of the classes.}
    \label{fig: adv_dist}
\end{figure}

\noindent{\textbf{Distribution of adversarial samples.}} To check how well the generated adversarial samples represent the remaining data, we visualize their distribution in relation to real samples. Using the SVHN and CIFAR-10 datasets with class 3 designated as the unlearning class, we generate adversarial samples to approximate the remaining classes. Figure \ref{fig: adv_dist} demonstrates that the adversarial samples closely align with real data distributions, particularly for classes adjacent to the unlearning class. In both datasets, we set class $3$ as the unlearning class. The generated samples closely overlap with the real samples, demonstrating a good approximation. This proximity arises from the reduced perturbation required to shift samples to nearby decision boundaries. In comparison, distant classes require larger perturbations. Nevertheless, as shown in the following, a limited number of adversarial samples per class suffices to estimate class-wise subspaces effectively.

\noindent{\textbf{Influence of adversary attack.}} We estimate the inaccessible remaining samples using generated adversary samples in our method. A pertinent question arises: \textit{How does the adversary attack affect the effectiveness of our method?} We conduct ablation studies using the same settings as in Section~\ref{sec: sc_unlearn}. These studies address this question by examining two aspects: 1) the influence of attack success rate (ASR), defined as the ratio of samples successfully identified as other classes; and 2) the choice of different attack methods.
 
\begin{figure}[tp]
    \centering
    \begin{subfigure}[b]{0.22\textwidth}
        \includegraphics[height=\textwidth]{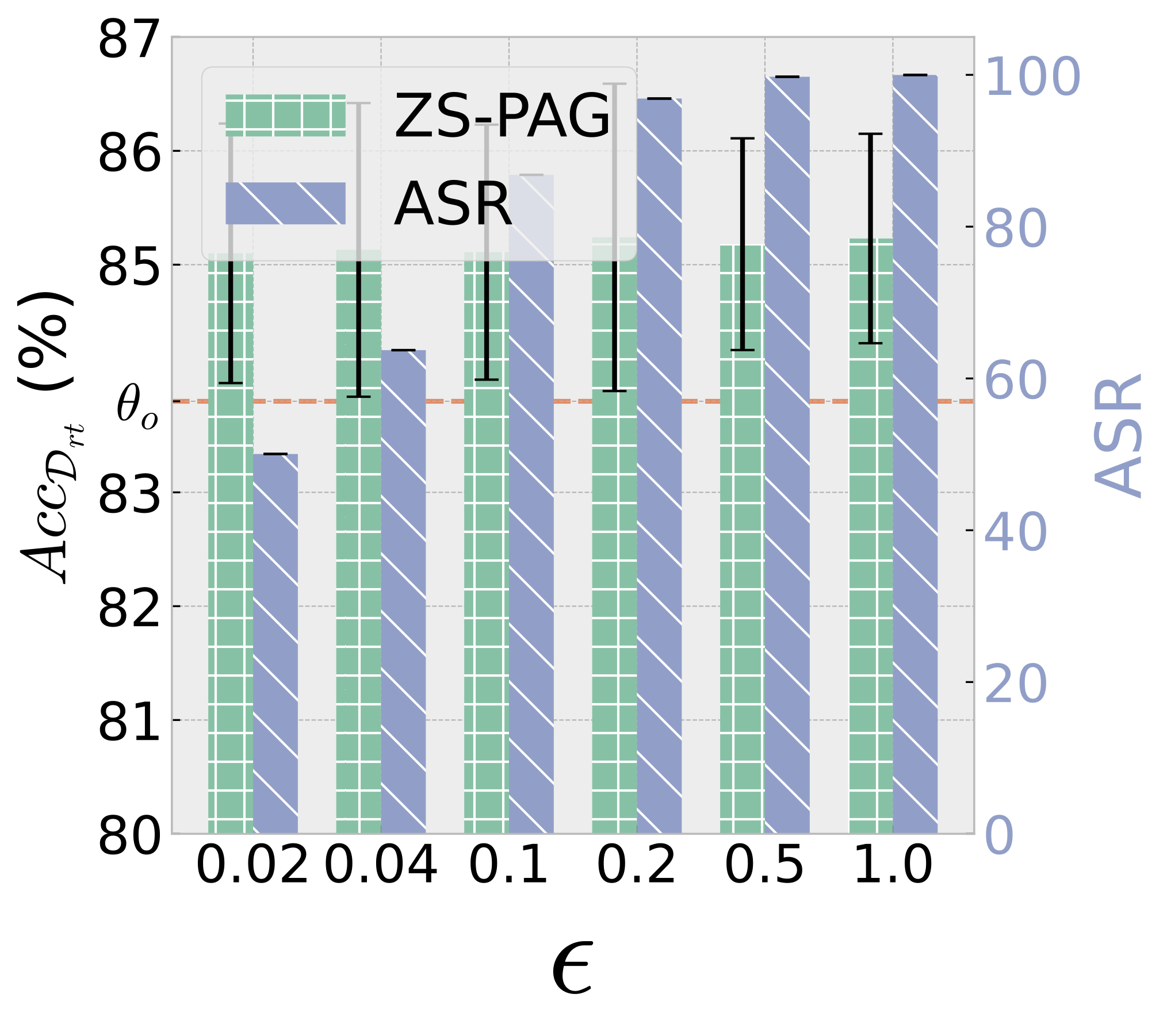}
    \caption{}
    \label{fig: noise bound}
    \end{subfigure}
    \hfill
    \begin{subfigure}[b]{0.22\textwidth}
        \includegraphics[height=\textwidth]{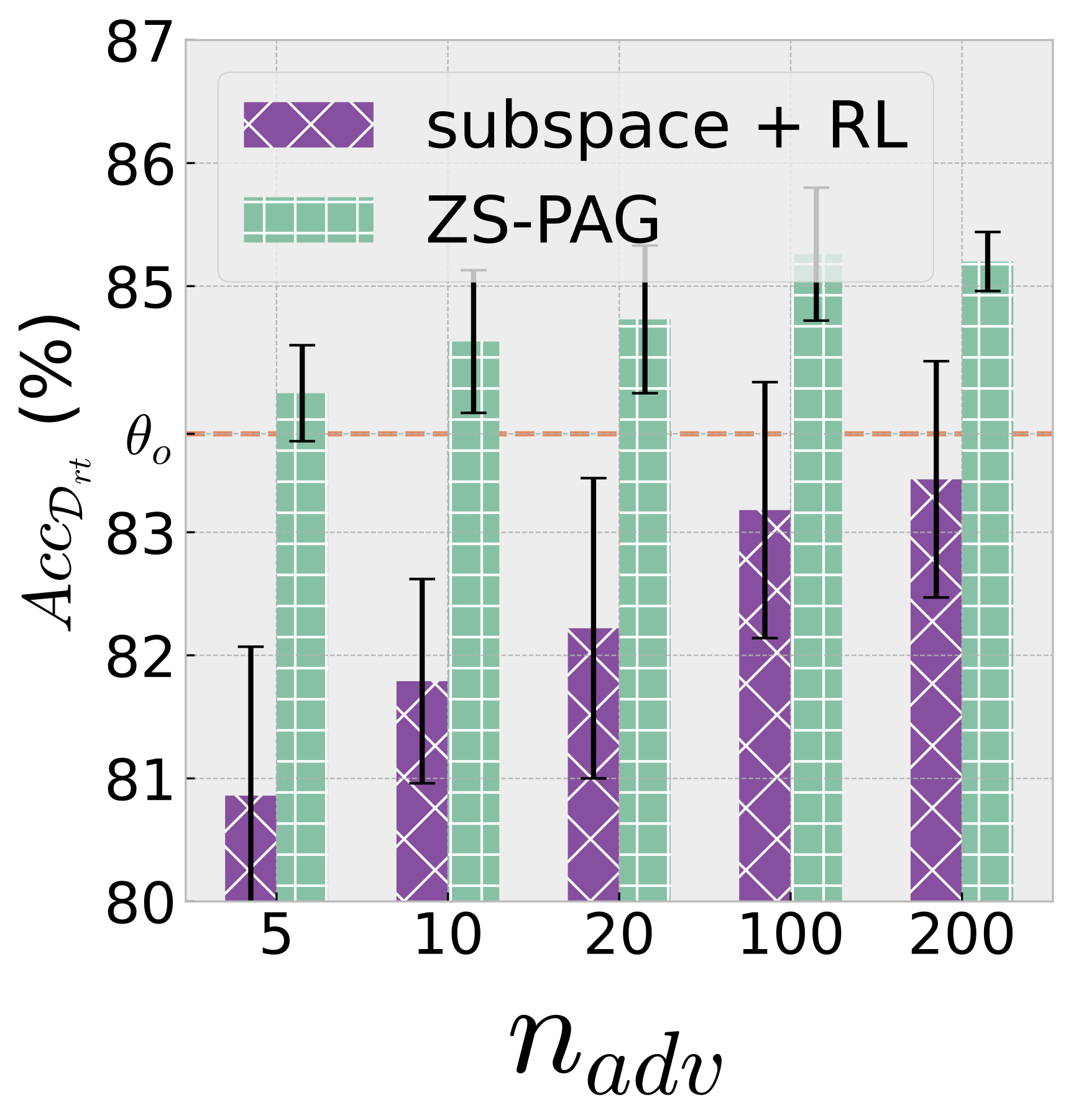}
        \caption{}
        \label{fig: num adv}
    \end{subfigure}
    \caption{Influence of adversary attack on CIFAR-$10$ dataset for single-class unlearning. (a) We use various levels of noise bound $\varepsilon$ when finding adversary samples. (b) We estimate the null space with different numbers of adversary samples.}
\end{figure}

\textit{Influence of ASR}: We generate adversarial samples with varying noise bound $\varepsilon$. As shown in Figure~\ref{fig: noise bound}, increasing $\varepsilon$ results in higher ASR due to larger perturbations. Notably, $Acc_{\mathcal{D}_{rt}}$ remains stable around $85\%$, indicating that ZS-PAG is robust to variations in ASR. Additional experiments, as depicted in Figure~\ref{fig: num adv}, confirm that as the number of generated adversarial samples per class $n_{\text{adv}}$ increases, the gap between the performance of the subspace-only method (\textit{Subspace + RL}) and the original model narrows, demonstrating that a sufficient number of adversarial samples enhances subspace estimation and mitigates over-unlearning. This observation aligns with the results shown in Figure \ref{fig: noise bound}, where we observe consistent performance across ASR levels ranging from $52\%$ to $99\%$. In the context of class unlearning, a substantial number of samples are available for adversarial attacks. Consequently, even a low ASR provides enough adversarial samples to meet our requirements. 

\textit{Influence of different choices of attack methods}: We compare four different attack methods, 1) PGD~\cite{madry2017towards} (the default method in this paper), 2) FGSM~\cite{fgsm}, 3) CW~\cite{cw}, and 4) DeepFool~\cite{deepfool}. Results in Table~\ref{tab: ablation_attck} show that all methods achieve comparable performance in terms of $Acc_{\mathcal{D}_{rt}}$, ranging from $84.46\%$ to $85.49\%$. Minor differences in $Acc_{\mathcal{D}_{ut}}$ are observed. These results demonstrate that ZS-PAG is robust to the choice of attack method. This robustness stems from the fact that ZS-PAG primarily focuses on probing the decision boundaries of the model rather than requiring high-quality adversarial samples. By focusing on the general properties of decision boundaries, our method accurately identifies the subspaces needed for unlearning, regardless of the attack method used.

\begin{table}[tp]
    \centering
    \begin{tabular}{c|cc}
    \toprule
    CIFAR-10/ViT & $Acc_{\mathcal{D}_{rt}}(\uparrow)$ & $Acc_{\mathcal{D}_{ut}}(\downarrow)$ \\
    \midrule
    Original & $ 83.80_{\pm 1.16} $ & $63.97_{\pm 0.46}$ \\
    Retrain & $ 86.57_{\pm 0.28} $ & $0.00_{\pm 0.00}$ \\
    \midrule
    PGD & $85.47_{\pm 1.49}$ & $1.40_{\pm 0.14}$ \\
    FGSM & $85.43_{\pm 1.06}$ & $2.00_{\pm 0.16}$ \\
    CW & $84.46_{\pm 1.11}$ & $0.90_{\pm 0.08}$ \\
    DeepFool & $85.49_{\pm 0.95}$ & $0.720_{\pm 0.11}$ \\
    \bottomrule
    \end{tabular} 
    \caption{\textbf{Influence of attack method.} We evaluate the impact of different adversarial attack methods on the unlearning performance.}
    \label{tab: ablation_attck}
\end{table}

\noindent{\textbf{Contribution of subspace projection.}} The efficacy of subspace projection in mitigating over-unlearning is analyzed through three experiments: (1) vanilla Random Labeling (\textit{RL}), (2) RL with a randomly generated subspace (\textit{Random Subspace+RL}), and (3) RL with the estimated subspace derived from adversarial samples (\textit{Estimate Subspace+RL}). As shown in Table \ref{tab: ablation_subspace}, \textit{RL} reduces  remaining class accuracy $Acc_{\mathcal{D}_{rt}}$ from from $95.52\%$ to $80.41\%$, causing significant over-unlearning. Employing a random subspace exacerbates this issue, lowering $Acc_{\mathcal{D}_{rt}}$ to $59.08\%$. In contrast, the estimated subspace effectively prevents over-unlearning, maintaining $Acc_{\mathcal{D}_{rt}}$ at $95.01\%$. These results suggest that the unlearning gradient must be approached with caution. Mitigating over-unlearning requires identifying a subspace significant to the remaining samples..

\begin{table}[tp]
    \centering
   \begin{tabular}{c|cc}
    \toprule
         SVHN/VGG & $Acc_{\mathcal{D}_{rt}}(\uparrow)$ &  $Acc_{\mathcal{D}_{ut}}(\downarrow)$ \\
         \midrule
          Original & $95.52_{\pm 0.12}$ & $91.30_{\pm 0.30}$  \\
          Retrain & $95.56_{\pm 0.23}$ & $0.00_{\pm 0.00}$  \\
          \midrule
          \textit{RL} & $80.41_{\pm 6.60}$ & $5.91_{\pm 2.80}$ \\
         Random Subspace + \textit{RL}  & $59.08_{\pm 4.71}$ & $1.15_{\pm 1.05}$ \\
         Estimate Subspace + \textit{RL} & $95.01_{\pm 0.33}$ & $1.46_{\pm 1.14}$ \\
         \bottomrule
    \end{tabular}
    \caption{
   \textbf{Contribution of Subspace Projection}. We conduct an ablation study to evaluate the contribution of subspace projection.}
    \label{tab: ablation_subspace}
\end{table}

\noindent{\textbf{Contribution of pseudo-labeling.}} We compare ZS-PAG with a subspace-only method that excludes pseudo-labeling. As illustrated in Figure 5b, the pseudo-labeling strategy consistently improves remaining class accuracy $Acc_{\mathcal{D}_{rt}}$, with performance gains ranging from $1.77\%$ to $3.27\%$, depending on the number of adversarial samples per class $n_{\text{adv}}$. This improvement underscores the effectiveness of influence-based pseudo-label optimization in enhancing the model’s overall performance.

\noindent{\textbf{Computational cost of ZS-PAG.}} We conduct the experiment on an NVIDIA RTX $4090$ GPU. We fix the unlearning epochs to 10 for a fair comparison. Tab.~\ref{tab:time_cost} indicates the running time and unlearning performance. ZS-PAG uses significantly less running time than Retrain and achieves the best unlearning performance, with the highest $Acc_{D_{rt}}$. Compared to the second-best method, Neggrad, ZS-PAG incurs less computing overhead and achieves better unlearning performance.
\begin{table}[htp]
    \centering
    \setlength{\tabcolsep}{3pt} 
    \resizebox{0.48\textwidth}{!}{
    \begin{tabular}{c|c|ccccc|c}
    \toprule
    & Retrain &  SalUn  & FT & BU & Neggrad  & Fisher  & ZS-PAG \\
    \midrule
    Time & 1h22m7s   & 5m13s & 5m14s & 5m16s & 10m12s & 21m34s & 8m28s\\
    \midrule
    $Acc_{D_{rt}}$& $75.36$  & $74.68$  & $74.68$ & $69.51$ & $75.15$ &  $50.03$ &  $\textbf{75.54}$ \\
    $Acc_{D_{ut}}$& $0.00$ & $1.67$ &$2.33$  & $3.00$ & $3.00$&  $0.00$ & $2.00$ \\
    \bottomrule
    \end{tabular}
    }
    \caption{\textbf{Computation overhead}.We unlearn 1 class from CIFAR100/ResNet18.}
    \label{tab:time_cost}
\end{table}

\section{Conclusion}
This paper presents ZS-PAG, a novel approach to zero-shot machine unlearning. ZS-PAG approximates the inaccessible remaining samples with adversary samples and confines the unlearning process within a specified subspace. This effectively prevents over-unlearning. Additionally, ZS-PAG integrates influence-based optimization techniques to enhance the unlearned model's performance further. Our theoretical analysis provides robust support for this approach, and empirical results convincingly demonstrate the superior performance of ZS-PAG over existing methods. Moreover, our method exhibits robustness across various model architectures, underscoring its effectiveness and adaptability.
\section*{Acknowledgments}
This research is partially supported by NSFC-FDCT under its Joint Scientific Research Project Fund (Grant No. 0051/2022/AFJ)

\bibliographystyle{named}
\bibliography{ijcai24}
\end{document}